\documentclass[letterpaper, 10 pt, conference]{ieeeconf}  % Comment this line out if you need a4paper

\IEEEoverridecommandlockouts                              % This command is only needed if 
\usepackage{graphics} % for pdf, bitmapped graphics files
\usepackage{epsfig} % for postscript graphics files
\usepackage{amsmath} % assumes amsmath package installed
\usepackage{amssymb}  % assumes amsmath package installed

\usepackage{multirow}

\usepackage{algorithm}
\usepackage{algorithmic}

\usepackage{subcaption}

\graphicspath{{images/}}

\newcommand{\etal}{\textit{et al}. }
\newcommand{\ie}{\textit{i}.\textit{e}. }
\newcommand{\eg}{\textit{e}.\textit{g}. }

\title{\LARGE \bf
Intend-Wait-\textit{Perceive}-Cross: Exploring the Effects of Perceptual Limitations on Pedestrian Decision-Making
}

\author{Iuliia Kotseruba$^{1}$ and Amir Rasouli$^{2}$% <-this % stops a space
\thanks{$^{1}$York University, {\tt\ yulia\_k@eecs.yorku.ca}. Work done while at Huawei.}%
\thanks{$^{2}$Huawei Technologies Canada, {\tt amir.rasouli@huawei.com}}%
}

\begin{document}

\maketitle
\begin{abstract}
Current research on pedestrian behavior understanding focuses on the dynamics of pedestrians and makes strong assumptions about their perceptual abilities. For instance, it is often presumed that pedestrians have omnidirectional view of the scene around them. In practice, human visual system has a number of limitations, such as restricted field of view (FoV) and range of sensing, which consequently affect decision-making and overall behavior of the pedestrians. By including explicit modeling of pedestrian perception, we can better understand its effect on their decision-making. To this end, we propose an agent-based pedestrian behavior model Intend-Wait-Perceive-Cross with three novel elements: field of vision, working memory, and scanning strategy, all motivated by findings from behavioral literature. Through extensive experimentation we investigate the effects of perceptual limitations on safe crossing decisions and demonstrate how they contribute to detectable changes in pedestrian behaviors.

\end{abstract}
\section{Introduction}
Accurately predicting pedestrian behaviors is important for intelligent driving due to inherent risks of interactions between traffic participants \cite{rasouli2019autonomous}. Current research relies on large naturalistic datasets (\eg Waymo \cite{sun2020scalability} and Argoverse \cite{chang2019argoverse}) and simulation environments (\eg CARLA \cite{dosovitskiy2017carla}) for modeling behaviors of road users. Strong assumptions about their perceptual abilities are often made; although often not explicitly stated, the agents are presumed to have an omnidirectional view of the scene around them (\eg cropped map centered around the agent \cite{salzmann2020trajectron}). Although this may be true for autonomous vehicles with extensive sensor suites and human drivers who have access to mirrors, pedestrians rely only on their vision with limited field of view (FoV) and sequentially scan the scene during decision-making. 

Top-down (or bird's eye) views common in the autonomous driving datasets provide information on the dynamics of the agents but not their awareness or decision-making process. The latter may be captured implicitly, however, such unobserved variables cannot be effectively learned as recent findings indicate \cite{ruan2022learning}. Likewise, training on synthetic data with similar characteristics would increase sim-real gap, therefore simulations would greatly benefit from integration of more psychologically-plausible elements \cite{markkula2022explaining}. 

In this paper, we address the issue of explicitly modeling perceptual limitations by extending our previous work Intend-Wait-Cross \cite{rasouli2022intend} -- a microscopic agent-based model of pedestrian crossing behavior motivated by psychological studies. In our experiments, we demonstrate how perception operates and how it contributes to detectable changes in pedestrian behavior.

\begin{figure}[!t]
\centering
\includegraphics[width=0.8\columnwidth]{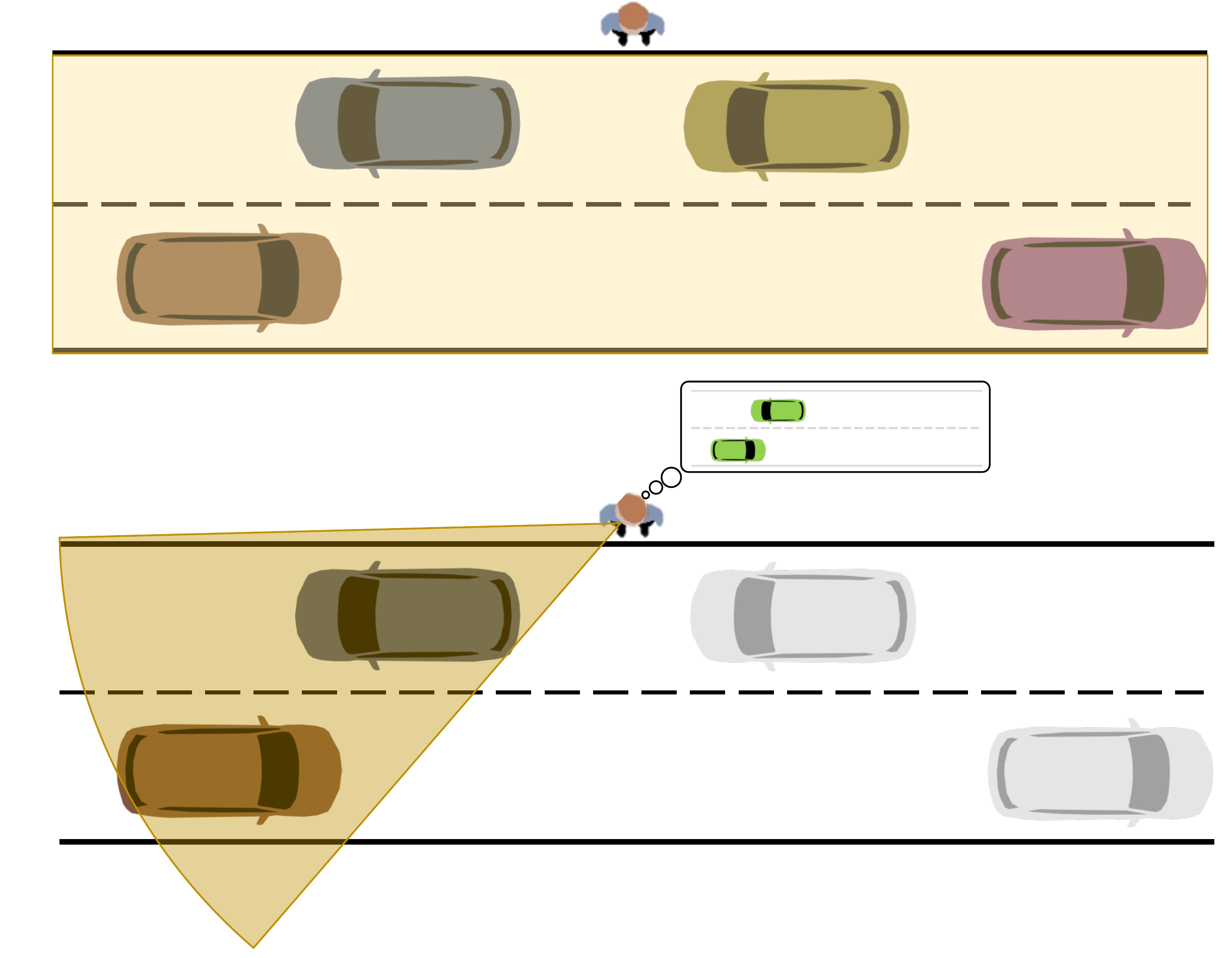}
\caption{Pedestrian's perception of the road. Top row: approach in \cite{rasouli2022intend} where pedestrian sees the entire road. Bottom row: the proposed perception model with limited field of view and working memory (shown as a thought bubble).}
\label{fig:first}
\vspace{-1em}
\end{figure}

\section{Related Work}
%\subsection{Behavior Simulation}
%\subsection{Human Perception modeling}
%\subsection{Crossing Behavior}
\noindent
\textbf{Pedestrian behavior simulation.} Microscopic agent-based simulations have successfully modeled a wide variety of phenomena emerging from interactions between multiple heterogeneous agents, particularly in the transportation domain \cite{bernhardt2007agent}. Due to risks involved, much of the literature focuses on jaywalking \cite{wang2021modeling, rasouli2022intend} and unsignalized crossing scenarios \cite{feliciani2017simulation, wang2021modeling, zhu2021novel, lu2016cellular, zhu2021interactions}, although some studies consider illegal crossings at signalized sites as well \cite{suh2013modeling}. 

\noindent
\textbf{Pedestrian perception models.} Unlike behavioral and demographic characteristics of pedestrians (\eg preferences, age, gender, walking speed), effects of perceptual limitations are less investigated and are incorporated only in some models. The most frequently modeled aspects are limited field of view and occlusions. For example, Lu \etal \cite{lu2016cellular} introduce driver's vision distance defined as a minimum distance needed for the vehicle to come to stop and also the moment at which the driver begins to pay attention to the pedestrian near the crossing area. The authors of \cite{zhu2021novel, zhu2021interactions} implement visual field of drivers and pedestrians as circular sectors with visibility radius of 100 m and 60 m, respectively. Without occlusions by other objects, pedestrians have a $180^{\circ}$ view. Drivers' visual field is adjusted depending on their speed to reflect tunneling, \eg $132^{\circ}$ for $<20$ km/h and $108^{\circ}$ for $>40$ km/h. 

Behavioral studies indicate that limited field of view is supported by representations in short-term memory that helps select and preserve spatial information across multiple viewpoints and enable decision-making and actions \cite{hayhoe2003visual, henderson2003human}. However, memory is notably absent from many agent simulations.
As an exception, simulation in \cite{seele2017integration} includes multi-purpose synthetic perception (including attention and FoV) and short-term memory, however, without specifics of implementation and limited proof-of-concept demonstrations in video game scenarios.

\noindent
\textbf{Learning crossing behavior.} In intelligent driving, planning relies on action and trajectory predictions of the other road users. While dynamic information is still the most dominant feature for prediction \cite{rudenko2020human}, taking into account different perceptual abilities of the traffic participants has demonstrable benefits. For example, \cite{hasan2018seeing} extends the energy-based model for pedestrian trajectory prediction with a frustum of attention ($30^{\circ}$ circular sector) towards where the pedestrian is heading. \cite{bi2020can} combines top-down view and simulated first person view with limited vision and attention to predict pedestrian trajectories. Similarly, \cite{kim2020pedestrian} uses pedestrian and vehicle perspectives to model their trajectories. 

Motivated by the benefits of explicit modeling of pedestrian perception in traffic simulation and prediction applications, we propose a pedestrian agent model with the following novel components (illustrated in Figure \ref{fig:first}): 1) a model of FoV with sensing range and viewing angle; 2) scanning strategy for observing the scene; and 3) working memory for temporary storage of observations.

\section{Pedestrian Behavior Model}
The proposed behavior model is based on the approach in \cite{rasouli2022intend}. This model considers various choices that pedestrians make, including their choice of transit, route, activity, speed, and next step. These choices are predominantly impacted by pedestrian characteristics: \textit{type} which is based on the age and determines walking speed; \textit{trait} -- aggressive, conservative, or average -- that affects walking speed and gap acceptance of the pedestrian; \textit{law obedience} -- violating, obedient, and average -- that determines likelihood of the pedestrian to jaywalk or go to the crosswalk; \textit{accepted gap} that the pedestrian considers safe to cross; \textit{crossing pattern} which including one-stage (cross when all lanes are safe) and rolling (cross when the immediate lane is safe); \textit{perceptual noise} -- error in the pedestrian estimation of the dynamic agents. In addition, the model of \cite{rasouli2022intend} proposes strategies for identifying relevant agents in the scene, estimating their TTCs, and changing behavior of the pedestrians due to the traffic conditions.

To better highlight the importance of modeling FoV and short-term memory, we focus on particular types of pedestrians. More specifically, we use law violating pedestrians as they rely on scanning their surroundings when they intend to cross as opposed to law abiding pedestrians at signalized intersections who mainly observe the signal when waiting to cross. In addition, we focus on one-stage crossing, where the pedestrian starts to cross only when sufficient gaps are available in all lanes.

Unlike the method in \cite{rasouli2022intend}, instead of allowing pedestrians to see the entire road in front of them at all times, we propose a psychologically-motivated perception model that limits the ability of the pedestrians to assess their surroundings. The details of the proposed approach are discussed in the following sections.

\section{Pedestrian Perception}

\begin{figure}[!t]
\centering
\includegraphics[width=0.75\columnwidth]{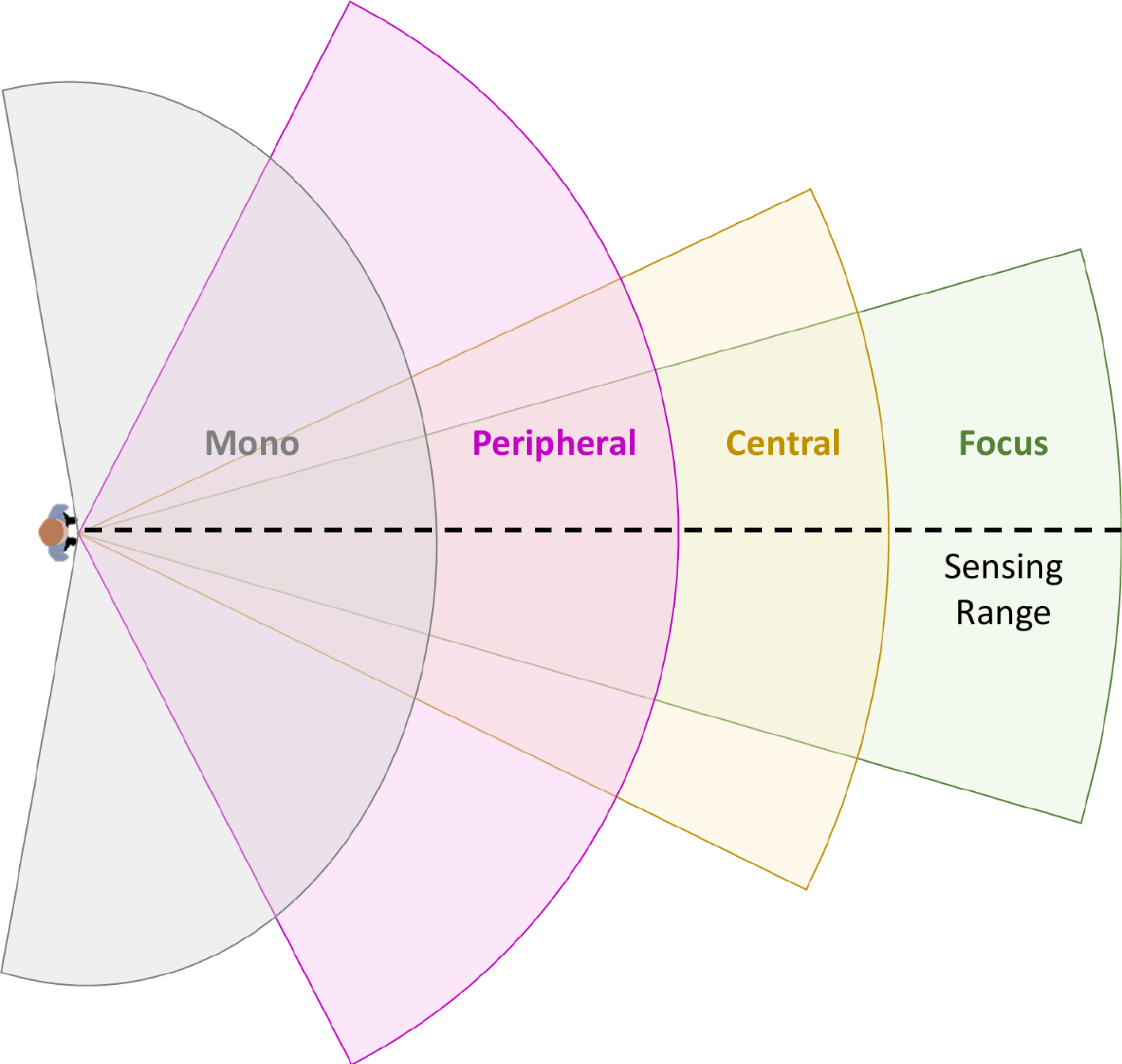}
\caption{Field of view divided into 4 regions that vary depending on sensing range and angle of view.}
\label{fig:FoV}
\vspace{-1em}
\end{figure}

\begin{figure*}[!th]
\centering
\includegraphics[width=0.9\textwidth]{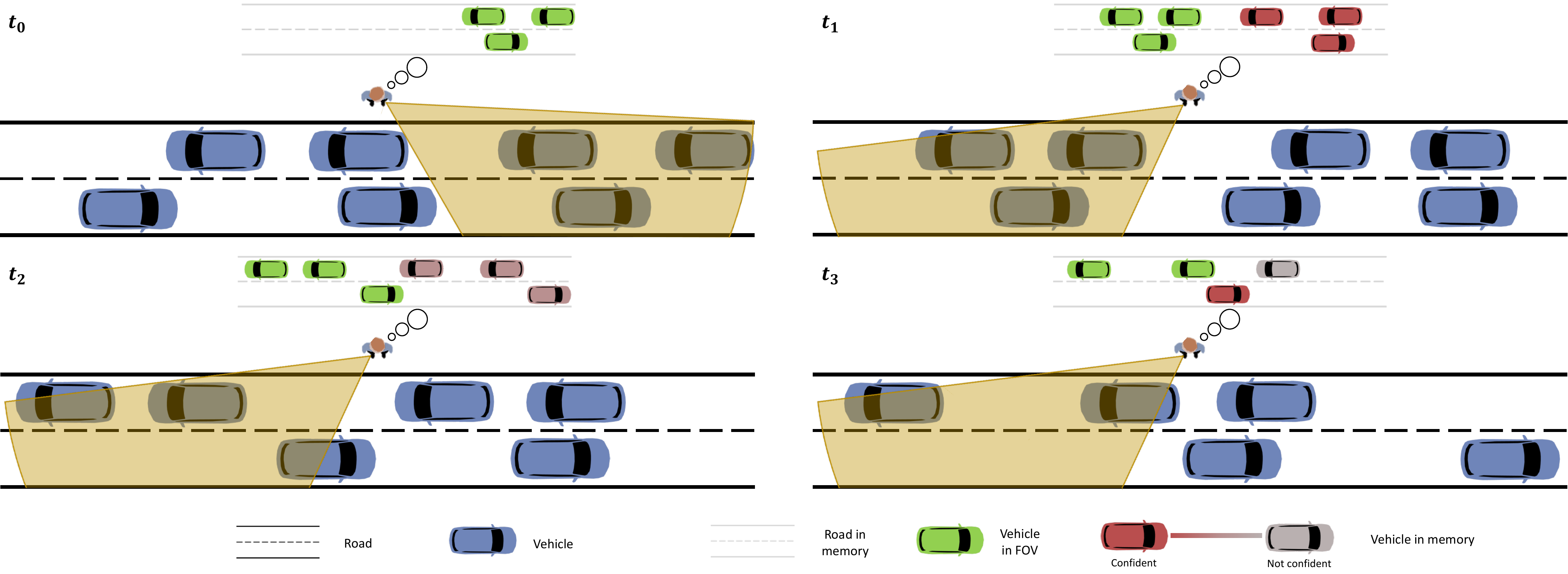}
\caption{Illustration of pedestrian observing the environment while registering and updating positions of the vehicles in memory. At $t_0$, the agent sees only the three vehicles in the FoV and registers them in memory (shown as a thought bubble). Since these vehicles are currently within the FoV, their true state (indicated by green color) is stored. At $t_1$ the pedestrian looks in the opposite direction, sees another three vehicles, and registers them. Status of the previously seen vehicles currently not in FoV does not truly reflect reality anymore due to the estimation error. As the time passes, the status of the vehicles not in FoV becomes more uncertain (shown in gradually fading red color). At $t_3$, one of the vehicles re-enters pedestrian's FoV and its state is reset to true value.}
\label{fig:mem}
\vspace{-1.5em}
\end{figure*}

In our model, pedestrian field of view is limited by sensing range and angle of view. The sensing range corresponds to the maximum distance within which pedestrian can observe objects and reason about their dynamics, \eg changes in their speed. Angle of view, as the name implies, defines  the angular extent of the scene observable by the pedestrian. 

To simulate foveation of human vision, \ie resolution that is higher in the center of the visual field than in the periphery \cite{osterberg1935topography}, we divide visual field into sectors with varying sensing range and angular extent. As illustrated in Figure \ref{fig:FoV}, FoV is composed of four circular sectors, namely,  \textit{focus}, \textit{central}, \textit{peripheral}, and \textit{mono}. Central \textit{focus} cone has the longest sensing range but the smallest angular extent, and \textit{mono}, which corresponds to the monocular vision, has the shortest sensing range and the largest angular extent.

In the proposed decision model, the vehicles are considered as seen when any part of them fall within FoV. To determine the visibility of a vehicle, we consider two reference points -- the center coordinates of the front and back bumpers of the vehicle -- as follows:

\begin{equation}
 \begin{aligned}
   & Veh_{seen} = \forall{veh^i} \wedge \forall{view^k} | 
    \theta_{veh^i} <= \frac{1}{2} view_{ang}^k \\
   & \wedge veh_{dist}^i <= view_{sr}^k, i \in 1,2,...,n \\
   & k \in \{\mathrm{focus, central, peripheral, mono}\},\\
    \end{aligned}
\end{equation}
\noindent where $ang$ and $sr$ are angular extent and sensing range of the FoV sector, respectively. The angle $\theta_{veh^i}$ is given by

\begin{gather*}
       \measuredangle{v^ip} = tan^{-1} \frac{veh^i_y - ped_y}{veh^i_x - ped_x} * 180/\pi\\
      \theta_v^i = (\measuredangle{v^ip} - \theta_p + 360)\mod 360\\
      \theta_p = \phi_{head} + \phi_{body}.
\end{gather*}

\noindent
Here $\measuredangle{v^ip}$ is the angle between $veh_i$ and the pedestrian $ped$, and $\phi_{head}$ and $\phi_{body}$ are the pedestrian's head and body angles respectively. Once a vehicle is seen, its information is registered in the working memory and can be used for making crossing decisions.

\section{Working Memory}

\subsection{Memory contents}
Given that pedestrians can observe only a part of the scene at a time, temporary storage (here referred to as working memory) is needed to aggregate observations across multiple views. In doing so, working memory maintains the states of the dynamic objects that are not currently seen and updates their status based on the beliefs of the pedestrian.

In the proposed memory model, every time a vehicle is seen, its id, dynamic characteristics, and the observation timestamps are registered in the memory. After that, depending on whether the vehicle is still being observed or not, its status is updated following an update rule. 

\subsection{Memory update}
To maintain a reasonable representation of the scene in the memory, its state should be updated at every time step. The status of the vehicles currently seen can be updated based on their true state in the world. However, the status of the vehicles that are registered in the memory but are not currently in the field of view (FoV) should be updated based on a belief system. An overview of the memory registration and update is reflected in Figure \ref{fig:mem}. 

In our model, positions of registered vehicles that are not currently in FoV are updated by applying a dynamic update rule to their last observed states (speed, position, acceleration) and the time has passed since the vehicles were last seen. The updating position of a vehicle $\widehat{veh}$ in memory is computed as follows:

\begin{gather}
    \widehat{veh}_{pos} = [veh_x + sin(\phi_{v})\cdot\widehat{d}, veh_y + cos(\phi_{v})\cdot\widehat{d}]
\end{gather}
\noindent where $\phi_{v}$ is the heading angle of the vehicle at the time of observation and $\widehat{d}$ is the distance that the pedestrian believes the vehicle has travelled since last observation. This distance is given by
\begin{gather*}
    \hat{d} = veh_s \cdot (t - t_{v\_seen})+ \frac{1}{2} \cdot veh_{accl}\cdot (t - t_{v\_seen})^2 
\end{gather*}
\noindent where $veh_s$ and $veh_{accl}$ are the vehicle's last seen speed and acceleration, and $t_{v\_seen}$ is the time the vehicle was last seen. 

From the above formulation one can see that the pedestrian estimation of the vehicle's state may not be accurate due to the changes in the environment. For example, while pedestrian is not looking, vehicles may slow down or accelerate in response to traffic signals or other vehicles' actions. Naturally, this error accumulates as long as the pedestrian looks away from the given vehicle, thus increasing the risk of crossing decision based on outdated information.

Vehicles registered in the memory but not in FoV can be observed again, \eg due to the pedestrian's head movement or if the vehicle moves into pedestrians' FoV. In such cases, status of vehicles is reset to their true state in the world.

\label{sec:scanning_strategy}
\begin{figure*}[!th]
\centering
\includegraphics[width=\textwidth]{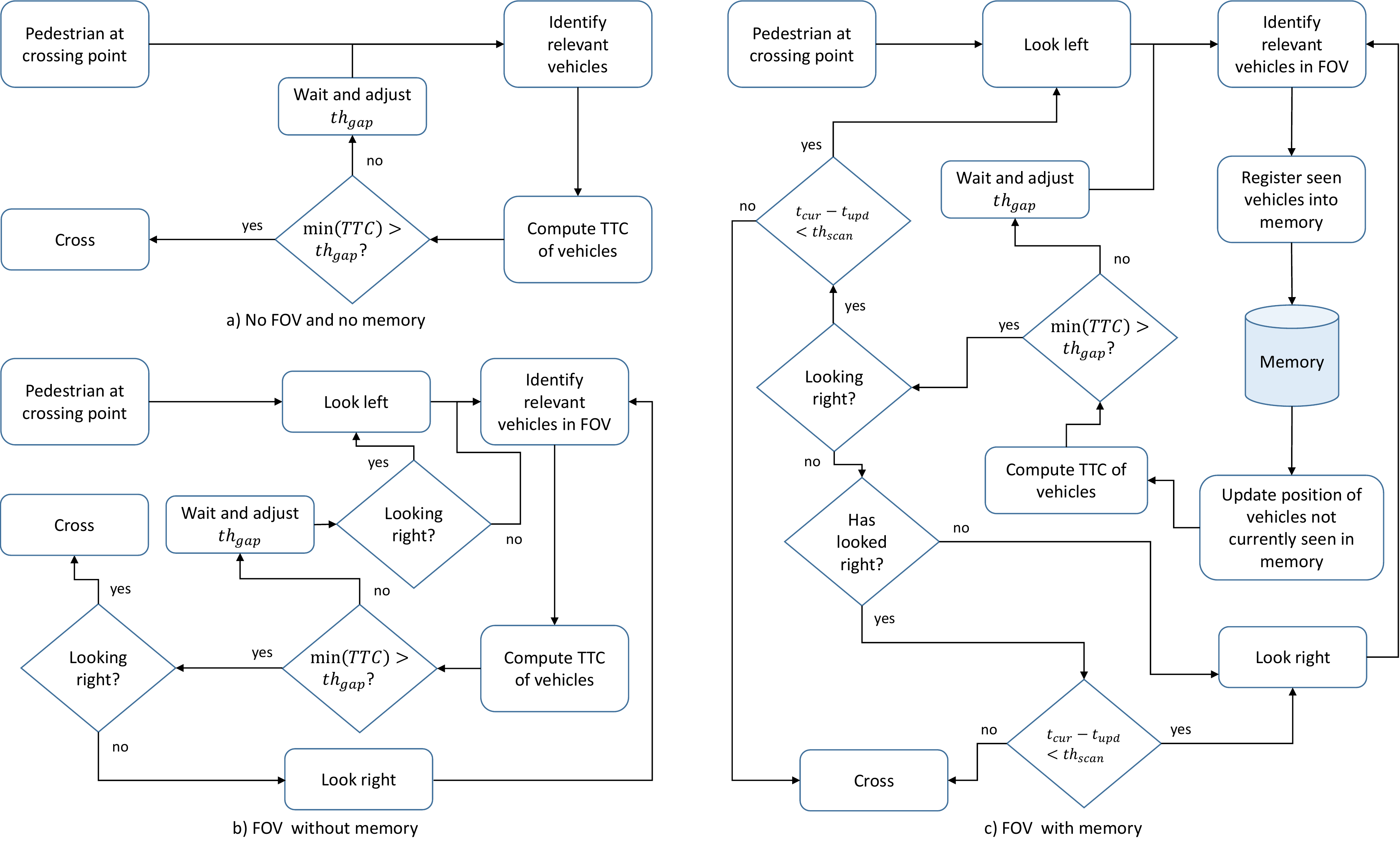}
\caption{An overview of different scanning strategies in a) the absence of FoV model and memory as in \cite{rasouli2022intend}, b) with only FoV model and c) the proposed model with FoV and working memory. }
\label{fig:diagram}
\vspace{-1em}
\end{figure*}

\subsection{Removal from memory}
In our memory model, the vehicles' information are stored and maintained as long as their current position is within the maximum sensing range of the pedestrian, \ie within the sensing range of the \textit{focus} sector. Once the vehicles are believed to leave or seen to leave maximum sensing range, their information is removed from the memory. Additionally, if vehicles become irrelevant (\eg passes the pedestrian) and no longer pose any risk, they are removed from the memory.

Finally, when the pedestrian finishes crossing, storing the state of the vehicles on the road becomes unnecessary and the contents of working memory is flushed out.

\section{Scanning Strategy}

Having a limited field of view, the pedestrian scans the environment sequentially. Thus, a strategy is needed to observe relevant parts of the scene in a timely manner. For the roads with right-hand traffic, a reasonable strategy is to look first to the left, and if the road is clear, then look to the right. We follow a similar approach to model the scanning behavior of pedestrians (see diagrams in Figure \ref{fig:diagram}).

The pedestrian starts by looking to the left until they find safe gap for potential crossing. Then they turn to the right and look for a safe gap on the other side. If the right side is safe, they start crossing while pointing their head towards the direction of the traffic on the current lane. 

In the absence of working memory, if the pedestrian does not find a safe gap on the right side, they have to immediately look to the left again as the status of the next immediate lane which is not currently in FoV is unknown to them. This results in very fast head movement and unrealistic behavior. 

Working memory allows the pedestrian to assess the right side while updating the status of the vehicles seen previously on the left. Hence, if a safe gap is identified on the right within a reasonable time, they can commence crossing without the need to look back.

As mentioned earlier, the pedestrian's estimation of the vehicles' states not in FoV can be erroneous, and the error increases as the time passes. Moreover, there is a chance that unobserved vehicles can get too close to the pedestrians. As a result, if a safe gap was not identified after the first scan, the pedestrian should look back and forth to maintain more accurate representation of the environment.

% Please add the following required packages to your document preamble:
% \usepackage{graphicx}
\begin{table*}[!th]
\centering
\caption{Effects of observation mode, memory, and perceptual noise on crossing and collisions. \textit{All} - omnidirectional perception with no memory as in \cite{rasouli2022intend}; \textit{FoV} - field of view; \textit{FoV+mem} uses \textit{dynamic} memory update (vehicle speed and position are calculated taking into account acceleration at the time of last update). Results for each mode are reported with and without perceptual noise.}
\resizebox{\textwidth}{!}{%
% \begin{tabular}{l|ccc|ccc|ccc}
%  & \multicolumn{3}{c|}{Wait time (s)} & \multicolumn{3}{c|}{Num. veh-person collisions} & \multicolumn{3}{c}{Memory error (m)} \\ \hline
% Observation mode & light & med & heavy & light & med & heavy & light & med & heavy \\ \hline
% Original & \textbf{2.8 (5.2)} & 3.6 (5.6) & \textbf{4.9 (8.9)} & \textbf{0.1 (0.3)} & 0.2 (0.4) & 0.1 (0.3) & - & - & - \\
% FoV & 3.9 (6.3) & 4.8 (7.4) & 5.9 (9.2) & 0.3 (0.5) & 0.2 (0.6) & 0.2 (0.4) & - & - & - \\
% FoV + memory (sim) & 3.9 (5.7) & 4.6 (7.0) & 5.9 (9.1) & 0.3 (0.5) & \textbf{0.1 (0.3)} & 0.4 (0.5) & - & - & - \\
% FoV + memory (static) & 3.5 (4.8) & \textbf{4.1 (6.2)} & 5.3 (8.6) & 0.5 (0.5) & 0.3 (0.7) & \textbf{0.1 (0.3)} & 7.1 (4.8) & 6.7 (4.7) & 7.0 (4.9) \\
% FoV + memory (constant) & 3.6 (5.1) & 4.6 (7.0) & 5.7 (8.9) & 0.3 (0.5) & 0.5 (0.8) & 0.4 (1.0) & 1.5 (1.7) & 1.6 (1.7) & \textbf{1.7 (1.7)} \\
% FoV + memory (dynamic) & 3.6 (5.0) & 4.6 (7.3) & 5.9 (9.0) & 0.3 (0.5) & 0.4 (0.7) & 0.6 (1.0) & \textbf{1.3 (1.9)} & \textbf{1.4 (1.8)} & \textbf{1.7 (2.3)}
% \end{tabular}%
\begin{tabular}{ll|lll|lll|lll}
                           &         & \multicolumn{3}{c|}{Wait time (s)}   & \multicolumn{3}{c|}{Min TTC (s)}  & \multicolumn{3}{c}{Num. veh-person collisions} \\ \cline{2-11} 
                           & Traffic & light     & med        & heavy       & light     & med       & heavy     & light          & med           & heavy         \\ \cline{2-11} 
\multirow{3}{*}{W/o noise} & all     & 4.7 (8.0) & 6.1 (9.5)  & 8.6 (11.8)  & 6.8 (4.5) & 6.3 (4.5) & 5.0 (4.3) & 0.0 (0.0)      & 0.3 (0.5)     & 0.0 (0.0)     \\
                           & FoV     & 5.7 (9.0) & 6.5 (9.8)  & 8.5 (12.2)  & 4.9 (4.4) & 4.7 (4.2) & 4.3 (4.2) & 0.3 (0.5)      & 0.3 (0.7)     & 0.2 (0.4)     \\
                           & FoV+mem & 5.3 (8.3) & 6.1 (9.5)  & 7.5 (10.9)  & 5.5 (4.5) & 5.2 (4.4) & 4.6 (4.3) & 0.1 (0.3)      & 0.3 (0.5)     & 0.2 (0.4)     \\ \cline{2-11} 
\multirow{3}{*}{W/ noise}  & all     & 5.7 (8.9) & 8.7 (12.9) & 11.0 (14.5) & 7.1 (4.5) & 6.1 (4.4) & 5.4 (4.3) & 0.5 (0.7)      & 1.0 (0.8)     & 0.4 (0.8)     \\
                           & FoV     & 5.7 (8.7) & 6.7 (10.4) & 7.9 (11.8)  & 5.4 (4.4) & 4.4 (4.2) & 4.6 (4.2) & 0.2 (0.4)      & 0.3 (0.5)     & 0.7 (0.7)     \\
                           & FoV+mem & 5.0 (7.7) & 6.4 (9.8)  & 7.6 (11.2)  & 5.8 (4.4) & 5.0 (4.4) & 5.1 (4.4) & 0.7 (0.7)      & 0.4 (0.5)     & 0.4 (0.5)    
\end{tabular}
}
\label{tab:obs_mode_vs_traffic_density}
\end{table*}

To achieve this behavior, we setup a threshold, $th_{scan}$ to induce the pedestrian to move their head to the opposite direction if no safe gap was found. Specifically, the pedestrian changes head direction if $t_{cur} - t_{upd} > th_{scan}$, where $t_{cur}$ is the current time step and $t_{upd}$ is the last time pedestrian looked in the opposite direction. $th_{scan}$ is determined as 
\begin{equation}
    th_{scan} = max(view_{sr}^k)/road_{max_s} - ped_{c\_gap}
\end{equation}
\noindent where $road_{max_s}$ is the speed limit and $ped_{c\_gap}$ is gap accepted by the pedestrian. 
%can either be set statically based on a fixed time period or dynamically to reflect the confidence of the pedestrian in the status of the vehicles not currently in FoV. An overview of the scanning strategy is shown in Figure \ref{fig:diagram}.

\begin{figure}[!th]
\centering
\begin{subfigure}{0.9\columnwidth}
\includegraphics[width=\columnwidth]{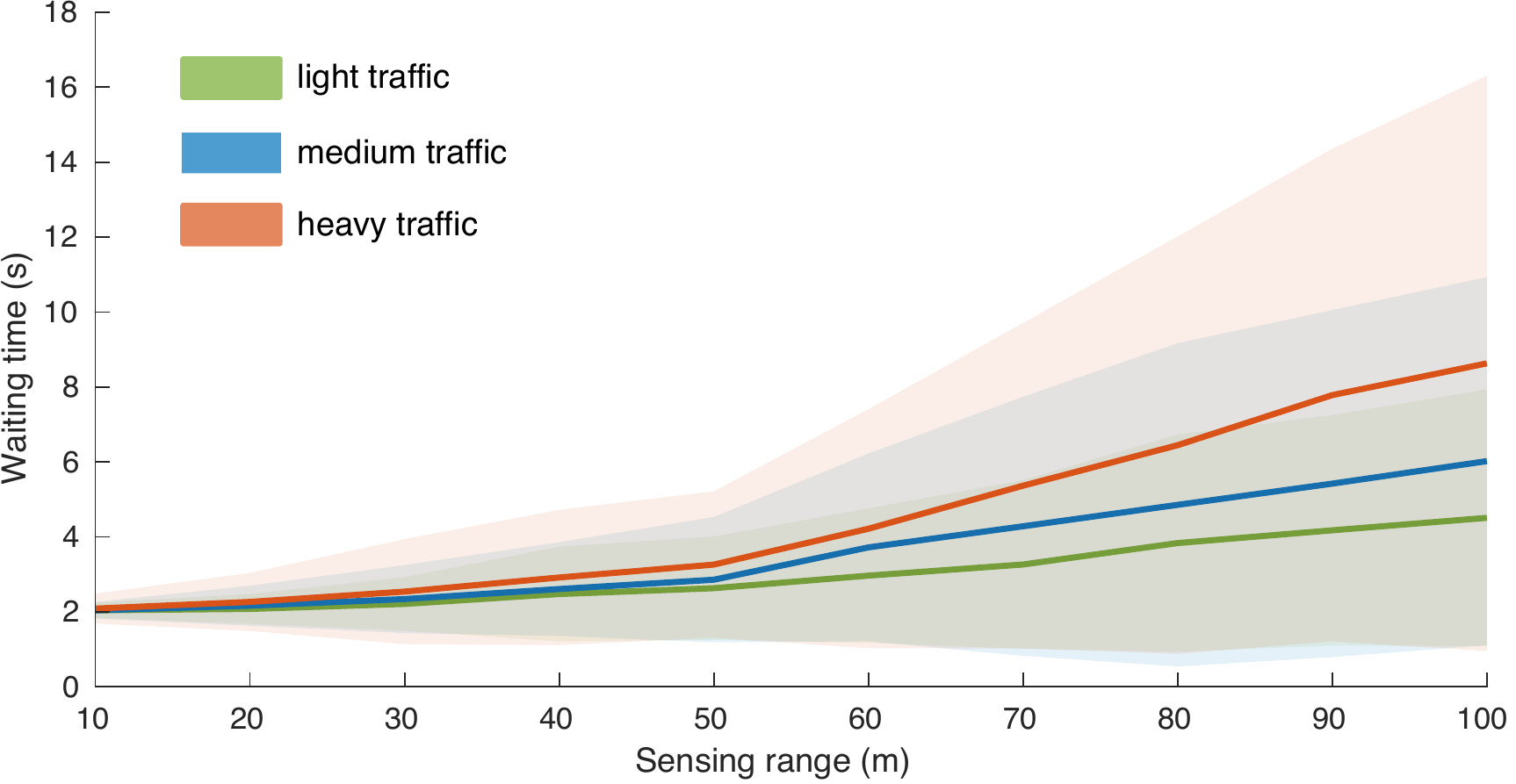}
\caption{}
\end{subfigure}
\centering
\begin{subfigure}{0.9\columnwidth}
\centering
\includegraphics[width=\columnwidth]{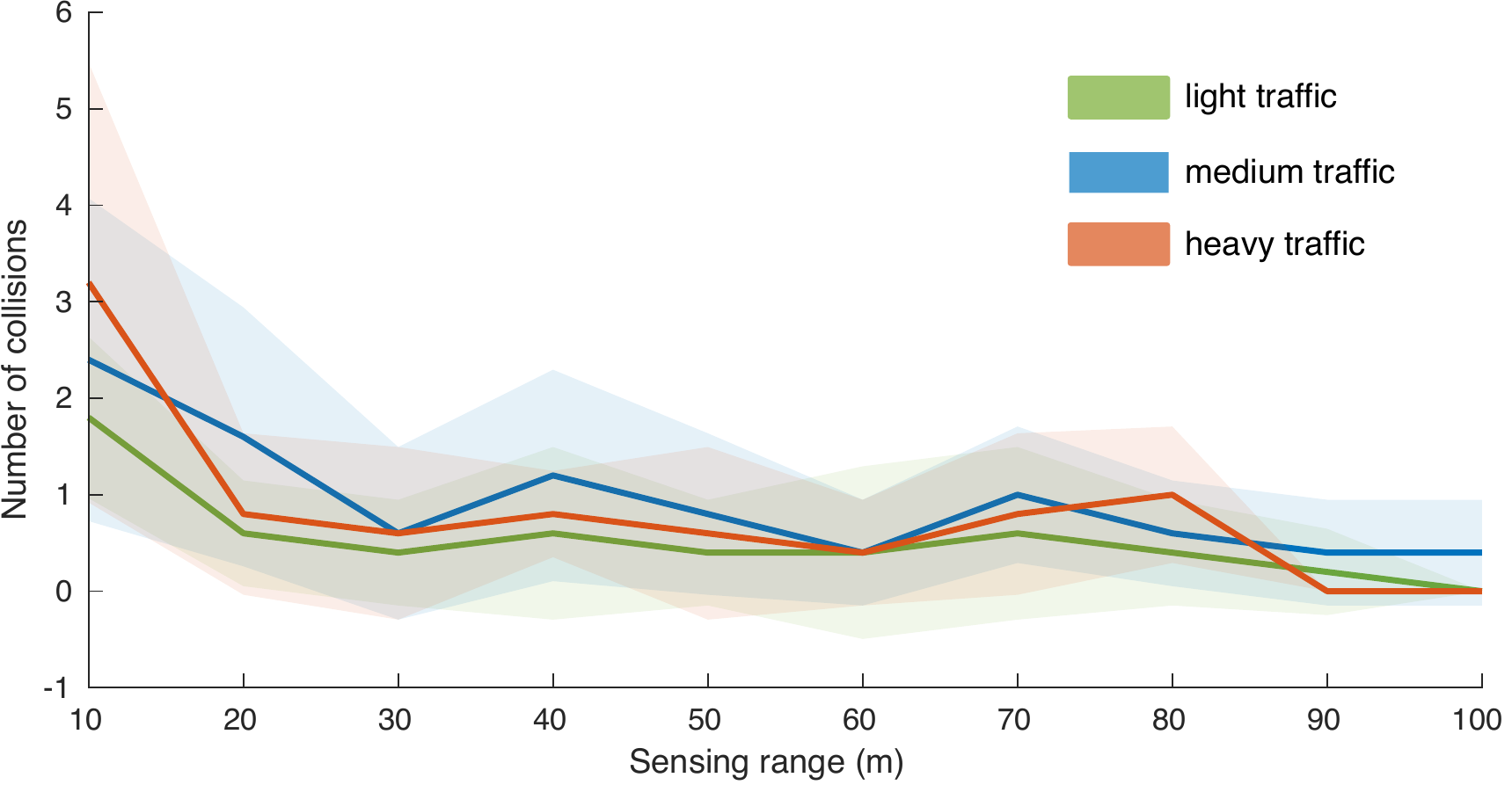}
\caption{}
\end{subfigure}
\caption{Effect of sensing range on (a) wait time and (b) number of collisions between vehicles and pedestrians. Solid lines and shaded regions show mean values and standard deviations.}
\label{fig:sensing_range}
\end{figure}

\section{Experiments}
We use a similar experimental setup as in \cite{rasouli2022intend} and the following default values for the perceptual components: sensing range - 80m, angular extents of view fields (focus - $30^{\circ}$, central - $60^{\circ}$, peripheral - $120^{\circ}$, mono - $190^{\circ}$) and dynamic memory update.

% max_sense_range 80
% degrade_ratio: 0.20
% view fields:
% focus 30
% central 60
% peripheral 120
% mono 190
% mem_decay auto
% scan_strategy env save
% memory update dynamic

We also set the law obedience of all pedestrians to \textit{law violating} since perceptual limitations are more safety-critical during jaywalking as opposed to crossing at designated crossings. Pedestrian type of all agents is set to \textit{adult} since we do not model differences in perception across ages.

\subsection{Effect of sensing range}
To examine the effect of sensing range, which determines how far the pedestrian can perceive the objects, we use the following setup: 270m long straight road segment with no traffic lights; sensing range varied from 10m to 100m with a step of 10m; light, medium, and heavy traffic is simulated by generating 1 vehicle every 6, 4, and 2s, respectively. Pedestrian memory is not used in these experiments and other parameters are set to default values. We run the simulation for all combinations of sensing range and traffic density 5 times for 500 steps with fixed random seeds.

% RERUN THE EXPERIMENT
%     x = [10, 20, 30, 40, 50, 60, 70, 80, 90, 100]
% veh_per_collisions
% 600
% y_mean [2.1 0.8 0.7 0.7 0.8 0.3 0.4 0.5 0.  0.1]
% y_std [1.66332999 0.91893658 0.8232726  0.67494856 0.91893658 0.67494856
%  0.51639778 0.52704628 0.         0.31622777]
% 900
% y_mean [2.  1.3 1.2 0.9 0.6 0.2 0.5 0.1 0.4 0.1]
% y_std [1.41421356 0.8232726  0.91893658 1.28668394 0.6992059  0.63245553
%  0.52704628 0.31622777 0.51639778 0.31622777]
% 1200
% y_mean [2.6 1.3 0.7 1.1 0.  0.6 0.3 0.2 0.2 0.3]
% y_std [1.17378779 0.67494856 0.9486833  0.99442893 0.         0.6992059
%  0.48304589 0.42163702 0.42163702 0.48304589]
% waittime
% 600
% y_mean [ 2.52475248  3.12298137  4.102657    5.03860523  5.80417755  6.71243523
%   7.64615385  8.71710526  9.56233422 10.03655352]
% y_std [ 2.06628238  3.55112358  5.4175444   6.87416313  8.24689497  8.62612692
%   9.66328749 10.07496131 11.22589318 11.32254253]
% 900
% y_mean [ 3.79808841  6.10861423  8.72171651 12.31241656 15.27461859 15.58423913
%  16.40960452 17.88493151 17.6755618  19.70140845]
% y_std [ 5.23341273  9.8489562  14.69571788 19.62178801 22.64170272 22.22093078
%  21.79703257 23.72517356 22.2007227  24.13626977]
% 1200
% y_mean [ 5.45899633 14.00142248 22.16467066 27.60506706 33.08071749 35.38880249
%  36.48303716 37.71428571 39.86942149 39.42262895]
% y_std [ 8.43972838 30.98016548 46.22803601 48.22562555 52.31107497 53.62775237
%  52.7823364  53.72307735 54.98435485 52.77386746]

Visualization of the results can be seen in Figure \ref{fig:sensing_range}. As expected, wait time grows and number of accidents decreases as both sensing range and traffic density increase. Visibility below 50m generally leads to very short wait times and very high number of collisions since pedestrians do not see approaching vehicles and start crossing. Above 50m, the mean waiting time continues to grow in all traffic conditions. Note that when pedestrians can see farther ahead, their wait time becomes more variable as it now depends more on the available gaps and risk taking characteristics of pedestrians (size of the gap they will accept). 

\subsection{Effect of field of view and memory}
Next, we test how the addition of FoV and memory affects behaviors of pedestrians. We set up a 4-way intersection with 2 lanes in each direction and vary traffic density (600, 900, and 1200 veh/h). We fix the sense range parameters at default values and test the following combinations of sense range and memory updates: \textit{all} -- the original Intent-Wait-Cross model with omnidirectional perception \cite{rasouli2022intend}, \textit{FoV} -- model with the limited field of view and no memory, and \textit{FoV+mem} -- model with both FoV and memory enabled. Additionally, we test the effects of perceptual noise on each condition. As before, each combination is run 5 times for 500 steps with fixed random seeds.

Table \ref{tab:obs_mode_vs_traffic_density} summarizes the results, showing the effects of noise, observation mode and memory updates on decision-making in terms of waiting time and accepted gap, as well as number of collisions between vehicles and pedestrians. 

Without perceptual noise, as expected, introduction of FoV increases wait time because pedestrians sequentially scan the road before crossing. Addition of the memory decreases the wait time by reducing the need for repeated scanning. The same trend can be observed in the case of noisy observations. However, the wait time in the \textit{all} condition is much higher due to the noise accumulation across the whole scene. The effect of noise is reduced by FoV and even more so by memory since updates rely on internal information rather than noisy observations.  

Limiting observation increases the chance of riskier decision as apparent in the reduced gap acceptance in the case of \textit{FoV}. Introduction of memory (\textit{FoV+mem}) increases the accepted gap, however it is still below the condition when the pedestrians see the entire scene (\textit{all}). 

As one would expect, with limited view, the number of collisions increases, which can be remedied to some extent by memory. But memory update uncertainty combined with perceptual noise and variable traffic leads to less predictable outcomes. In the light traffic conditions, vehicles are more spaced out and have more freedom to move around. This results in more variable behavior, which, combined with noisy internal representation, increases chance of collisions. In congested traffic, vehicles behave more uniformly and more predictably, therefore memory reduces the average number of collisions.

% The addition of FoV leads to shorter wait times across all conditions and smaller accepted gaps since pedestrians do not have a complete view of the road. Addition of memory counteracts limited range, however, in both cases pedestrians tend to take riskier decisions, as confirmed by the growing number of collisions with vehicles. The addition of perceptual noise leads to more variable patterns of behaviors. Agents with omnidirectional perception tend to be more conservative, take longer to cross, and make more mistakes due to errors in gap estimation. Interaction between perceptual noise, FoV and memory leads to more complex behavior patterns. [EXPLAIN]

% \subsection{Effect of scanning strategy}
% We test two variations of scanning strategy: a strategy described in Section \ref{sec:scanning_strategy}, referred to as \textit{left safe} and \textit{env safe}, where memory is not reset after the initial left-right scan. The experimental setup is identical to the ones described in the previous subsection, \ie 4-way signalized intersection with 2 lanes in each direction and light, medium, and heavy traffic density.

\subsection{Perceptual limitations and crossing behavior}

\begin{table}[]
\caption{Effects of perceptual limitations before and during crossing}
\label{tab:distraction}
\centering
\resizebox{0.7\columnwidth}{!}{%

% \begin{tabular}{lcc|ccc}
% \multicolumn{1}{c}{\begin{tabular}[c]{@{}c@{}}Obs.\\ mode\end{tabular}} &
%   \begin{tabular}[c]{@{}c@{}}Cross. \\ strategy\end{tabular} &
%   \begin{tabular}[c]{@{}c@{}}Cross \\ check\end{tabular} &
%   \begin{tabular}[c]{@{}c@{}}Wait \\ time (s)\end{tabular} &
%   Collisions &
%   \begin{tabular}[c]{@{}c@{}}Head \\ turns\end{tabular} \\ \hline
% FoV     & rolling   & yes & 6.4 (11.5)  & 1.2 (1.6) & 1.0 (0.0) \\
% FoV     & one-stage & yes & 13.1 (18.3) & 0.8 (1.3) & 4.2 (5.2) \\
% FoV     & one-stage & no  & 14.4 (17.7) & 0.6 (0.5) & 4.1 (4.0) \\
% FoV+mem & one-stage & yes & 12.2 (16.4) & 0.2 (0.4) & 4.3 (6.4) \\
% FoV+mem & one-stage & no  & 11.8 (15.6) & 0.4 (0.5) & 3.7 (4.9)
% \end{tabular}%
% }

% \begin{tabular}{lcc|clcc}
% \multicolumn{1}{c}{\begin{tabular}[c]{@{}c@{}}Obs.\\ mode\end{tabular}} &
%   \begin{tabular}[c]{@{}c@{}}Cross. \\ strategy\end{tabular} &
%   \begin{tabular}[c]{@{}c@{}}Cross \\ check\end{tabular} &
%   \begin{tabular}[c]{@{}c@{}}Wait \\ time (s)\end{tabular} &
%   TTC (s) &
%   Collisions &
%   \begin{tabular}[c]{@{}c@{}}Head \\ turns\end{tabular} \\ \hline
% FoV     & OS & - & 13.1 (18.3) & 6.0 (5.0) & 0.8 (1.3) & 4.2 (5.2) \\
% FoV+mem & OS & - & 11.8 (15.6) & 6.3 (4.8) & 0.4 (0.5) & 3.7 (4.9) 
% %FoV     & OS & + & 14.4 (17.7) & 5.2 (4.8) & 0.6 (0.5) & 4.2 (5.2) \\
% %FoV+mem & OS & + & 12.2 (16.4) & 6.6 (4.8) & 0.2 (0.4) & 3.7 (4.9)
% \end{tabular}%
\begin{tabular}{c|cc}
Obs. mode & Head turns & Collisions \\ \hline
FoV & 4.2 (5.2) & 0.8 (1.3) \\
FoV+mem & 3.7 (4.9) & 0.4 (0.5)
\end{tabular}%
}
\vspace{-2em}
\end{table}

Here, we look into effects of perceptual limitations prior and during the crossing. We use the same setup is as in the previous experiment, with medium traffic density and perceptual parameters set to default values. The results are summarized in Table \ref{tab:distraction}.

\textbf{Head movement} is often seen as a cue for intention to cross \cite{rasouli2017agreeing}. As expected, without internal memory representation, the number of head turns increases because pedestrians repeatedly observe the environment to find a safe gap to cross. This confirms the validity of the proposed perceptual and memory components.

\textbf{Distraction during crossing.} In the experiments so far, pedestrians continuously monitored the traffic post crossing decision and reacted to imminent risk by stopping mid-road to wait for another gap. Here, pedestrians complete the crossing without checking traffic, which doubles the number of accidents. Without memory, the initial crossing decision is made based on the final observation, whereas with memory the status of previously observed vehicles is also considered, helping to make a better decision.

\section{Discussion and future work}
In this paper, we argued that for realistic modeling of pedestrian behaviors their perceptual limitations should be taken into account. We proposed the Wait-Intend-Perceive-Cross model with three psychologically-motivated components: limited FoV, working memory, and scanning strategy. %Where possible, we used the literature to set values for parameters, otherwise plausible values are used. 

We conducted experiments to validate the proposed components and demonstrated their significant effects on pedestrian behavior, notably waiting times, head movements, gap acceptance, and collision risk. This suggests that perceptual limitations should be considered when modeling pedestrian behaviors for planning in the context of intelligent driving systems. In turn, a new approach to data collection is needed to record information, such as pedestrian demographics, head movement, and other relevant visual information. Although we made effort to maximize the realism of the proposed model, lack of data limited our ability to validate it and calibrate its parameters.

%We hope that this work can inform future data collection efforts, as many of the parameters of our model can be observed and measured, \eg wait times and changes in head orientation.

Extensions to be considered for future work include modeling perceptual limitations of drivers, effects of occlusions, and testing the impact of realistic perceptual models on the performance of prediction algorithms.
% working memory limit and updates
% currently, working memory size is unlimited

\bibliographystyle{IEEEtran}
\bibliography{references}
\end{document}